\documentclass[10pt,twocolumn,letterpaper]{article}

\usepackage[pagenumbers]{cvpr} 

\definecolor{cvprblue}{rgb}{0.21,0.49,0.74}
\usepackage[pagebackref,breaklinks,colorlinks,allcolors=cvprblue]{hyperref}

\usepackage{tabularx}
\usepackage{multirow}
\usepackage{arydshln}
\usepackage{booktabs}
\usepackage{amssymb}
\usepackage{xcolor}
\usepackage{enumitem}
\usepackage{graphicx}
\usepackage{lipsum}
\usepackage{amsmath}
\usepackage{amssymb}
\usepackage{booktabs}
\usepackage{mathtools}
\usepackage{tikz}
\usepackage{tikz-3dplot}
\usepackage{algorithm}
\usepackage{algorithmicx}
\usepackage{algpseudocode}
\usepackage{txfonts}
\usepackage{multicol}
\usetikzlibrary{calc,3d, positioning, arrows.meta, angles, quotes}
\usepackage[pagebackref,breaklinks,colorlinks]{hyperref}
\usepackage{listings}

\definecolor{codegray}{rgb}{0.95,0.95,0.95}  
\usepackage{siunitx}
\usepackage[capitalize]{cleveref}
\crefname{section}{Sec.}{Secs.}
\Crefname{section}{Section}{Sections}
\Crefname{table}{Table}{Tables}
\crefname{table}{Tab.}{Tabs.}

\newcommand{\secref}[1]{Section~\ref{#1}}

\newcommand{\figref}[1]{Fig.~\ref{#1}}
\newcommand{\eqnref}[1]{Eq.~\ref{#1}}
\newcommand{\tabref}[1]{Table~\ref{#1}}

\newcommand{\figreftwo}[2]{Figs.~\ref{#1} and~\ref{#2}}

\def\paperID{15681} 
\def\confName{CVPR}
\def\confYear{2026}

\makeatletter
\newcommand{\thickhline}{%
    \noalign{\ifnum0=`}\fi\hrule \@height 1.2pt \futurelet
    \reserved@a\@xthickhline}
\def\@xthickhline{\ifx\reserved@a\thickhline
    \vskip\doublerulesep
    \vskip-\thickarrayrulewidth
    \fi
    \ifnum0=`{\fi}}
\makeatother

\newlength{\thickarrayrulewidth}
\title{Realizing Fully-Integrated, Low-Power, Event-Based\\Pupil Tracking with Neuromorphic Hardware}

\author{Federico Paredes-Vall\'es$^{1}$\hspace{16pt}Yoshitaka Miyatani$^{2}$\hspace{16pt}Kirk Y. W. Scheper$^1$\vspace{5pt}\\
   $^{1}$Sony Advanced Visual Sensing AG, Schlieren, Switzerland\\
   $^{2}$Sony Semiconductor Solutions Corporation, Atsugi, Japan\\\
}

\begin{document}
\maketitle
\begin{abstract}
Eye tracking is fundamental to numerous applications, yet achieving robust, high-frequency tracking with ultra-low power consumption remains challenging for wearable platforms. While event-based vision sensors offer microsecond resolution and sparse data streams, they have lacked fully integrated, low-power processing solutions capable of real-time inference. In this work, we present the first battery-powered, wearable pupil-center-tracking system with complete on-device integration, combining event-based sensing and neuromorphic processing on the commercially available Speck2f system-on-chip with lightweight coordinate decoding on a low-power microcontroller. Our solution features a novel uncertainty-quantifying spiking neural network with gated temporal decoding, optimized for strict memory and bandwidth constraints, complemented by systematic deployment mechanisms that bridge the reality gap. We validate our system on a new multi-user dataset and demonstrate a wearable prototype with dual neuromorphic devices achieving robust binocular pupil tracking at 100 Hz with an average power consumption below 5 mW per eye. Our work demonstrates that end-to-end neuromorphic computing enables practical, always-on eye tracking for next-generation energy-efficient wearable systems.
\end{abstract}

\vspace{-12pt}
\section{Introduction}
Eye tracking is fundamental to a wide range of applications, from enhancing user interfaces and enabling medical diagnostics to advancing AR/VR, assistive technologies, and neuroscience \cite{Holmqvist2011,valtakari2021eye,Adhanom2023,Sqalli2023,pauszek2023introduction,FischerJanzen2024}. These applications demand precise, rapid estimation of eye position to ensure natural interaction and effective analysis of visual attention \cite{Andersson2010,Leube2017,Nystrom2021}. However, achieving robust, high-frequency tracking with low power consumption remains a significant challenge, particularly for wearable platforms \cite{Santini2018,Zhu2024,carminati2025energy}.

Traditional eye-tracking systems typically rely on frame-based cameras and computationally intensive algorithms. While capable of high spatial and moderate temporal resolution, these systems suffer from substantial power requirements and latency \cite{stein2021comparison,Feng2024_BlissCam,carminati2025energy}. The continuous capture and processing of dense frames leads to inefficiencies unsuitable for lightweight, battery-powered devices. Furthermore, they are vulnerable to motion blur and limited dynamic range, which degrade performance during rapid eye movements or under challenging lighting conditions \cite{kothari2020gaze,angelopoulos2020event,hooge2023robust}.

\begin{figure}[t]
	\centering
	\includegraphics[width=0.95\linewidth]{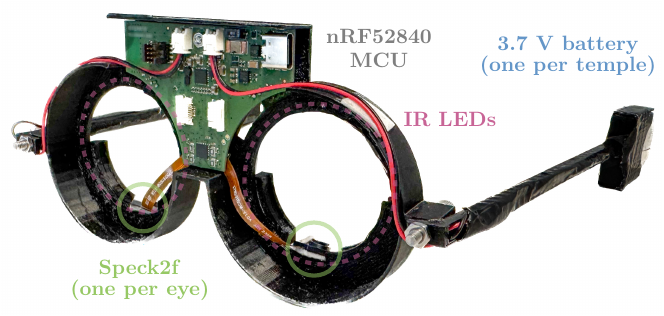}
    \vskip 3pt
    \includegraphics[width=\linewidth]{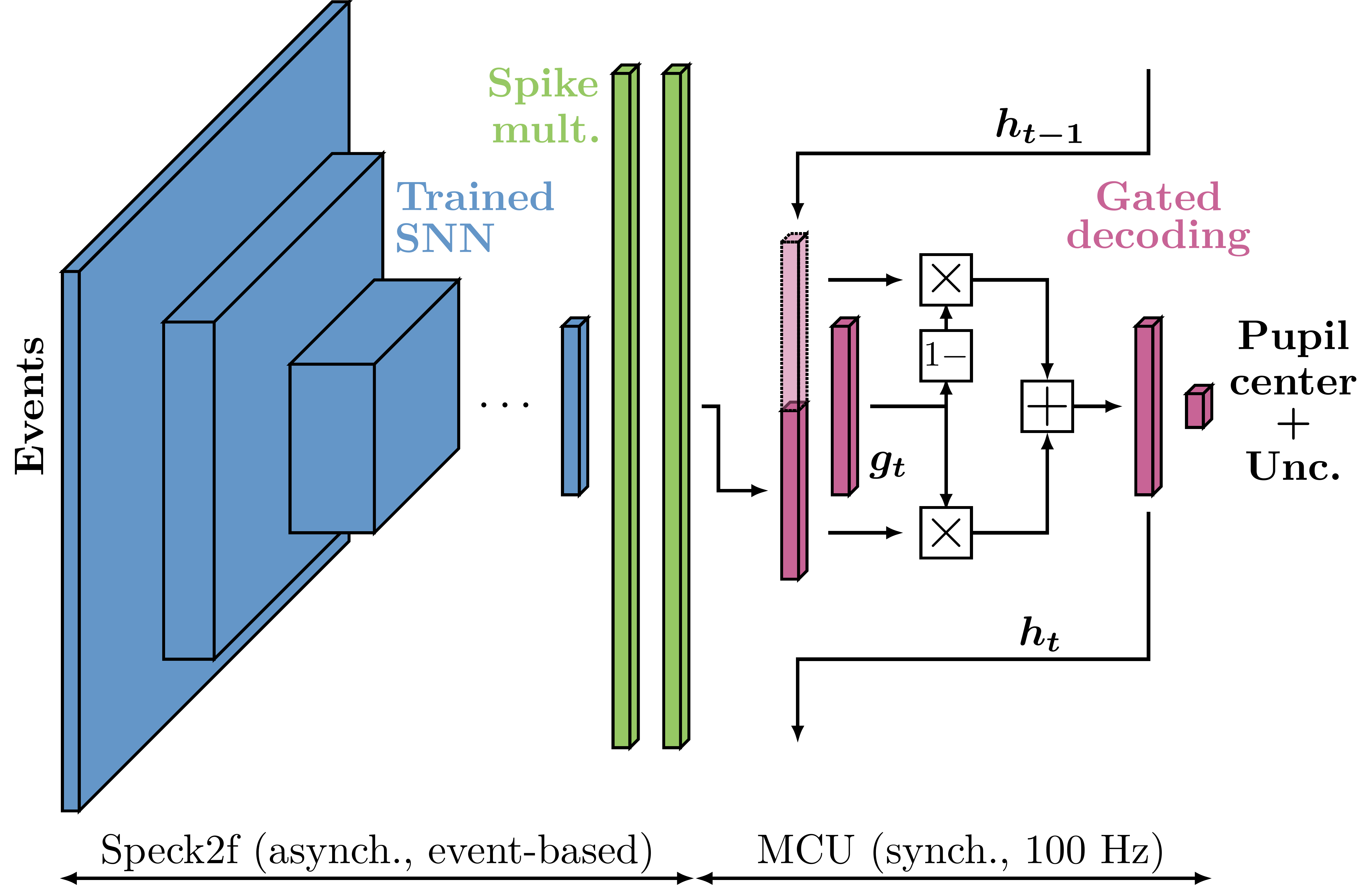}
    \caption{
    \textit{Top}: Prototype of our fully-integrated, battery-powered pupil-center-tracking system based on neuromorphic technology, featuring two Speck2f devices, IR LEDs, and an nRF52840 MCU. \textit{Bottom}: Our on-chip SNN processes input events asynchronously, while an MCU runs our off-chip decoding at 100~Hz to yield pupil-center and uncertainty estimates.}
	\label{fig:combined_glasses_arch}
\end{figure}

Event-based vision sensors (EVS) have recently emerged as a promising alternative. Unlike conventional image sensors, an EVS asynchronously detects per-pixel changes in log-brightness at microsecond resolution, producing sparse data streams that encode dynamic scene information \cite{Gallego2019_EventBasedVisionSurvey}. This enables low-latency, high-frequency tracking and robustness to motion blur, while dramatically reducing power consumption \cite{chen20233et,EventBasedEyeTracking2025Challenge,Zhang2024_SEE}. However, the asynchronous and sparse nature of event data demands fundamentally different processing approaches, yet most existing methods rely on dense event representations and processing, limiting efficiency and scalability \cite{Rebecq2019_BringingCVtoEventCameras,paredes2021back,chen20233et,Zhang2024_SEE,wu2024lightweight}.

Achieving high-frequency, low-power eye tracking with EVS requires not only efficient sensing but also tightly integrated, event-driven processing. Neuromorphic computing offers a promising avenue through spiking neural networks (SNNs), which process information via discrete spikes, potentially enabling real-time computation at ultra-low power~\cite{davies2018loihi, richter2023speck, yao2024spike}. The sparse and asynchronous nature of both event data and spike-based processing makes neuromorphic systems naturally suited for EVS applications. However, existing solutions targeting deployment on dedicated neuromorphic processors rely on power-hungry development kits, offload computation, or fail to fully exploit hardware capabilities~\cite{paredes2019unsupervised, hagenaars2021self, paredes2023taming, paredes2024fully, caccavella2024low, hines2025compact}.

In this work, we introduce the first fully integrated, battery-powered, wearable pupil-center-tracking system based on neuromorphic hardware. Leveraging the Speck2f system-on-chip (SoC)~\cite{richter2023speck, yao2024spike}, which integrates an EVS with dedicated SNN cores, we deploy a compact SNN that regresses pupil coordinates while respecting strict hardware constraints. By coupling event-based sensing, spike-based processing, and a low-power microcontroller (MCU) for efficient output decoding, we achieve robust, high-frequency tracking at very low power. Our contributions are:

\setlist[itemize]{
    leftmargin=20pt,
    itemsep=0pt,
    topsep=0pt,
    parsep=0pt
}
\begin{itemize}
    \item A novel uncertainty-quantifying SNN with gated temporal decoding and deployment pipeline optimized for the Speck2f's strict resource and readout constraints.
    \item The first fully integrated, battery-powered, wearable system for real-time, binocular pupil-center tracking combining event-based sensing and neuromorphic processing on dual Speck2f SoCs.
    \item Comprehensive validation on a new multi-user dataset and on a wearable prototype, demonstrating robust, high-frequency operation and low system-level power consumption.
\end{itemize}

\section{Related work}
Event-based eye tracking has gained significant attention due to the unique advantages of the EVS, such as low latency and high dynamic range. Early approaches include classical rule-based pipelines~\cite{angelopoulos2020event, stoffregen2022event}, while recent work focuses on lightweight deep learning models~\cite{chen20233et, wang2024mambapupil, ding2024facet, pei2024lightweight, zhao2023ev, Zhang2024_SEE}. Notably, prior work demonstrated gaze tracking at kilohertz rates enabled by the microsecond resolution of the EVS~\cite{angelopoulos2020event, stoffregen2022event}, and emphasized the importance of capturing temporal dynamics via advanced recurrent architectures~\cite{wang2024mambapupil}. Despite these advances, most solutions require substantial computational resources or offline processing, limiting their suitability for wearable applications.

To address these limitations, neuromorphic computing has emerged as a promising paradigm for event-based processing. Several works have explored SNNs for event-based eye tracking~\cite{jiang2024eye, groenen2025gaze}, but despite achieving high accuracy, these solutions remain undeployed on actual neuromorphic hardware. The sole exception is Bonazzi~\textit{et al.}~\cite{bonazzi2024retina}, who also tackled pupil-center tracking by deploying an SNN on the Speck2f SoC~\cite{richter2023speck, yao2024spike}.
However, their system relied on a power-hungry development kit and lacked support for continuous, real-time tracking due to network state resets between forward passes during training. Fully integrated, real-time, low-power event-based eye tracking on neuromorphic hardware therefore remains an open challenge.

Beyond eye tracking, other event-based solutions have been developed for neuromorphic hardware targeting applications like face detection \cite{caccavella2024low}, optical flow estimation \cite{dupeyroux2021neuromorphic, paredes2024fully}, and large language models~\cite{abreu2025neuromorphic}, among many others \cite{liu2019live, orchard2021efficient, ning2023neuromorphic, shrestha2024efficient, hines2025compact, angelo2025wandering}. While these solutions demonstrate the potential for energy-efficient, real-time inference, most implementations rely on external computers for computation or interfacing, which undermines overall system-level power savings and limits practical deployment in battery-powered operation scenarios.

In this work, we address these limitations by achieving full system integration: all sensing and processing are performed on-device within a single, wearable platform that combines event-based sensing, sparse asynchronous processing, and a low-power MCU interface. Uniquely, our system deploys a novel SNN on two Speck2f SoCs, one per eye, enabling simultaneous, real-time pupil-center tracking with low system-level power consumption, hence demonstrating the practical viability of neuromorphic hardware for always-on wearable applications.

\subsection{SynSense Speck2f}\label{sec:speck}
The core hardware in this work is the SynSense\footnote{See: \url{https://www.synsense.ai/}} Speck2f, a commercially available digital SoC that integrates a $128 \times 128$ EVS array with nine asynchronous convolutional SNN cores on a single die~\cite{richter2023speck, yao2024spike}. This tight integration enables efficient, low-power, event-driven sensing and processing, making it well suited for always-on applications.

A key feature of the Speck2f is its fully asynchronous, event-driven processing. Unlike other digital neuromorphic architectures that use global synchronization to trigger neuron updates~\cite{davies2018loihi, davies2021advancing}, the Speck2f updates its spiking neurons only when new input events arrive. This design allows each core to process spikes independently and immediately, routing internal spikes between cores via a network-on-chip without global coordination. Each core implements all neurons for a given convolutional layer, and processes input spikes in a first-in, first-out manner, performing (i) 2D convolution, (ii) spiking neuron update, and (iii) optional sum pooling for each spike.

\begin{figure}[t]
	\centering
	\includegraphics[width=0.765\linewidth]{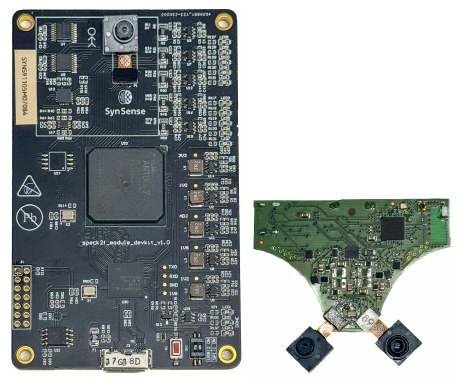}
    \vspace{-5pt}
    \caption{Speck2f support boards used in this work. \textit{Left}: Speck2f development kit, featuring a Xilinx Artix-7 FPGA. \textit{Right}: our custom PCB designed to accommodate two Speck2f units and featuring an nRF52840 MCU. The proportions of the boards are maintained to provide an accurate comparison of their sizes.}
	\label{fig:boards}
\end{figure}

Regarding the spiking model, the Speck2f supports leaky integrate-and-fire neurons~\cite{richter2023speck, yao2024spike, speck_specsheet, samna_docs}, where each neuron integrates incoming spikes with a linear leakage term and emits at most one output spike per update if the state exceeds a predefined threshold. In practice, leakage is applied synchronously via an external clock, which is at odds with the chip’s otherwise asynchronous design. As a result, most prior work omits the leakage mechanism~\cite{richter2023speck, caccavella2024low, brancaoptical, yao2024spike}, treating neurons as integrate-and-fire (IF) units without leakage. This makes neuron states dependent solely on the order of spikes, not their precise timing.

Each core operates under strict resource constraints, including 8-bit weights, 16-bit internal states, and limited bandwidth. The bandwidth of a core refers to the maximum number of synaptic operations\footnote{A synaptic operation includes all computational steps required to process an incoming spike.} per second (SOPs) that can be processed without introducing latency. Most cores support up to 30M SOPs, while the first core, which is directly connected to the EVS, supports up to 100M SOPs~\cite{yao2024spike}. 

For interfacing, SynSense mainly provides a development kit based on a Xilinx Artix-7 FPGA (see \figref{fig:boards}, left). All published work on this SoC to date has used this kit~\cite{richter2023speck, caccavella2024low, brancaoptical, bonazzi2024retina, hines2025compact}, which enables high-speed communication with a host computer via SynSense's proprietary protocol but incurs high static power consumption (approx.\ 0.6 W), far exceeding the Speck2f's own static power draw (approx.\ 0.5 mW), preventing low system-level power consumption.

The only alternative is a serial peripheral interface (SPI) 
that communicates directly with the Speck2f's dedicated readout core~\cite{richter2023speck, yao2024spike}. This core, driven by an external slow clock (SCLK), supports up to 16 output neurons (with the first remaining silent), each with a dedicated processing engine for temporal averaging of spike counts over configurable windows. The readout core can report either the index and activity of the most active neuron or the activity of a specific neuron. Notably, most existing approaches employ more than 15 active output neurons~\cite{caccavella2024low, brancaoptical, bonazzi2024retina, yao2024spike, hines2025compact}, exceeding the hardware limits of the on-chip readout and rendering them incompatible with this native, low-power interface.

\section{Method}
The goal of this work is to develop the first fully integrated, low-power system for continuous, real-world pupil-center tracking using event-based sensing and processing. We design a specialized network architecture that fits the strict memory constraints of the Speck2f, along with optimized training and deployment pipelines. We demonstrate our solution by running it in parallel on two Speck2f devices connected to a custom circuit board with a low-power MCU, achieving efficient battery-powered operation integrated in a wearable prototype. The following sections detail each component of our framework.

\subsection{Input format for off-chip training}\label{sec:input}
For an ideal EVS~\cite{Gallego2019_EventBasedVisionSurvey}, each event is generated at a pixel location and time when the log-brightness change since the last event at that pixel exceeds a polarity-specific threshold.

In Speck2f, events are sent directly from the sensing to the processing cores as they occur, without synchronization or accumulation~\cite{richter2023speck, yao2024spike}. To avoid simulating microsecond-level event processing during training, we discretize event streams into non-overlapping time windows of duration $\Delta t$, as in \cite{hagenaars2021self, paredes2023taming, caccavella2024low, paredes2024fully, yao2024spike}. Within each window, multiple events at the same pixel are aggregated into two-channel count images (one per polarity), providing a dense event representation. In the limit as $\Delta t \to 0$, this discretized approach approximates fully asynchronous event processing.

\subsection{Network architecture and neuron model}\label{sec:arch}
A schematic of our model is shown in \figref{fig:combined_glasses_arch} (bottom). The core SNN employs $L = 7$ of the nine available Speck2f processing cores to form a feed-forward convolutional network comprising just 46.2k parameters. This network maps the two-channel $128 \times 128$ EVS input to 15 active output neurons, thereby matching the SPI-based readout limit. Each layer applies a strided $3 \times 3$ convolution, with channel counts per layer of $\{4, 12, 18, 27, 40, 60, 15\}$, chosen to fit core memory and feature map constraints~\cite{richter2023speck, yao2024spike}. Similar to the linear leakage (see \secref{sec:speck}), bias terms are omitted to eliminate the overhead associated with synchronous neuron updates~\cite{richter2023speck, yao2024spike, speck_specsheet, samna_docs}.

For the activation function, we use IF neurons optimized for compatibility with the Speck2f:
\begin{align}
\tilde{v}_{i}[t] &= v_{i}[t\!-\!1] - s_{i}^{\text{out}}[t\!-\!1]v_{\text{th}} + \sum_j w_{ij}s_{ij}^{\text{in}}[t] \label{eq:voltage_raw} \\
v_i[t] &= \max(v_{\text{min}},\, \tilde{v}_i[t]) \label{eq:voltage_clip} \\
s_{i}^{\text{out}}[t] &=
\begin{cases} 
1, & \text{if } v_i[t] > v_{\text{th}} \text{ and on-chip}\\ 
\left\lfloor v_i[t]/v_{\text{th}} \right\rfloor, & \text{if } v_i[t] > v_{\text{th}} \text{ and off-chip}\\ 
0, & \text{otherwise}
\end{cases} \label{eq:spikes}
\end{align}

Here, \( v_{i}[t] \) denotes the internal state, commonly referred to as \textit{voltage}, of the \( i \)th neuron. This state evolves over time in response to weighted input spikes $\boldsymbol{s}_{i}^{\text{in}}$ and the output spikes \( s_{i}^{\text{out}} \) it generates, clipped below by $v_{\text{min}}$ = $-10$ to mimic on-chip state quantization. On-chip, a neuron emits a single spike if its voltage exceeds the threshold $v_{\text{th}}$ = 1. Off-chip, we model the neuron to emit multiple spikes per timestep proportional to $v_i[t]/v_{\text{th}}$, reducing the reality gap between training with discrete time windows $\Delta t$ and deployment on asynchronous event-based hardware~\cite{caccavella2024low, yao2024spike}.

Additionally, we implement a soft reset~\cite{bellec2018long}: after each output spike, the voltage is reduced by $s_{i}^{\text{out}}v_{\text{th}}$, as supported by Speck2f \cite{richter2023speck,yao2024spike}. To ensure correct behavior, we constrain synaptic weights to remain below $v_{\text{th}}$ via projected gradient descent. Otherwise, the state may remain above \( v_{\text{th}} \) after an output spike, potentially increasing the reality gap. For training, we use surrogate gradients (periodic version of the inverse tangent derivative~\cite{weidel2021wavesense, caccavella2024low}) due to the non-differentiable spike generation in \eqnref{eq:spikes}, and, unlike~\cite{bonazzi2024retina}, reset neuron states only after each sequence, thus supporting continuous operation.

\subsection{Gated decoding: Regression from spikes}\label{sec:decoding}
To derive continuous pupil-center coordinates from the 15 output spiking neurons, we propose a lightweight, stateful off-chip decoding (see \figref{fig:combined_glasses_arch}, bottom) suitable for low-power MCUs. At each timestep, the output spiking activity $\boldsymbol{x}_{L,t}$ is concatenated with a hidden state of equal length $\boldsymbol{h}_{t-1}$. This vector passes through a fully connected layer (with learnable weights $\boldsymbol{W}_g$ and biases $\boldsymbol{b}_g$) with sigmoid activation $\sigma(\cdot)$ to compute per-neuron gating values $\boldsymbol{g}_t$, which control the update of the hidden state:
\begin{align}
    \boldsymbol{g}_t &= \sigma(\boldsymbol{W}_g [\boldsymbol{x}_{L,t}, \boldsymbol{h}_{t-1}] + \boldsymbol{b}_g)\\
    \boldsymbol{h}_t &= \boldsymbol{g}_t \odot \boldsymbol{x}_{L,t} + (1 - \boldsymbol{g}_t) \odot \boldsymbol{h}_{t-1}
\end{align}

The updated hidden state is min-max normalized and passed through a final linear layer to predict pupil-center coordinates (and aleatoric uncertainty, see \secref{sec:training}):
\begin{align}
    \boldsymbol{h}_t^{\text{norm}} &= \frac{\boldsymbol{h}_t - \min(\boldsymbol{h}_t)}{\max(\boldsymbol{h}_t) - \min(\boldsymbol{h}_t)}\\
    [\hat{x}, \hat{y}]^\top &= \sigma(\boldsymbol{W}_{xy} \boldsymbol{h}_t^{\text{norm}} + \boldsymbol{b}_{xy})\\
    \boldsymbol{\hat{u}} &= \boldsymbol{W}_{u} \boldsymbol{h}_t^{\text{norm}} + \boldsymbol{b}_{u}
\end{align}

Unlike direct \cite{hagenaars2021self, caccavella2024low, paredes2024fully} or temporally-weighted spike decoding \cite{bonazzi2024retina}, our gating mechanism adaptively integrates temporal information, improving robustness to variable pupil speeds and occlusions while remaining efficient (approx.\ 1.1k floating-point operations). We avoid more complex recurrent units to minimize off-chip computation.

\subsection{Training details}\label{sec:training}
We train our network to regress pupil-center coordinates from event-camera data using a loss function that combines uncertainty-aware tracking and activity regularization:
\begin{align}\label{eq:loss}
\mathcal{L}(\boldsymbol{x}_0, \theta) = \mathcal{L}_{\text{track.}}(\boldsymbol{x}_0, \theta) + \beta \mathcal{L}_{\text{reg.}}(\boldsymbol{x}_0, \theta)
\end{align}
where $\boldsymbol{x}_0$ and $\theta$ denote the input to the SNN and the model parameters, respectively, while $\beta=100$ is a scaling factor.

\textbf{Uncertainty-aware tracking:} 
We use the loss from Kendall et al.~\cite{kendall2017uncertainties}, which extends mean squared error of the predicted pupil-center location with a learned log variance $\hat{\boldsymbol{u}}:=\log\hat{\boldsymbol{\sigma}}^2$ to model aleatoric uncertainty:
\begin{equation}\label{eq:loss_track}
\mathcal{L}_{\text{track.}}(\boldsymbol{x}_0, \theta) = \frac{1}{2} \exp(-\hat{\boldsymbol{u}}) \|\boldsymbol{y} - \hat{\boldsymbol{y}}\|^2 + \frac{1}{2} \hat{\boldsymbol{u}}
\end{equation}
where $\boldsymbol{y}$ and $\hat{\boldsymbol{y}}$ are the ground-truth and predicted pupil-center locations, respectively.

This uncertainty term serves as both a dynamic discount factor and a regularizer, encouraging reliable predictions and preventing unbounded uncertainty. It improves robustness to occlusions (e.g., during blinking) and low-motion scenarios (i.e., low signal-to-noise ratio due to limited input events), and removes the need for costly ground-truth observability labels in real-world data.

\textbf{Activity regularization:}
To ensure real-time operation within the Speck2f's processing capacity, activity regularization is needed. Prior work targeting deployment on this SoC has employed either direct spike count penalization~\cite{caccavella2024low} or target-based optimization~\cite{bonazzi2024retina}, but both have drawbacks: the former is sensitive to the balancing factor $\beta$ in \eqnref{eq:loss} and ignores hardware constraints, while the latter can force unnecessary activity during periods of low input event rates. Instead, our approach leverages Speck2f-specific bandwidth constraints, defined in terms of SOPs (see \secref{sec:speck}), to match the capacity of the processing cores. We penalize internal activity only when these hardware-derived thresholds are exceeded, minimizing the reality gap and preserving on-chip accuracy. This allows the model to converge to the best possible tracking solution that respects these constraints.

For each layer $l$, SOPs are computed from the input activity and connectivity within a time window $\Delta t$:
\begin{align}
\text{SOPs}_l = \frac{\boldsymbol{x}_{l-1}\hspace{2pt} \boldsymbol{m}_{l-1, l}}{\Delta t}
\end{align}
where $\boldsymbol{m}_{l-1, l}$ is the binary connectivity matrix between layers $l-1$ and $l$.

The activity regularization term is then defined as:
\begin{align}\label{eq:loss_reg}
\mathcal{L}_{\text{reg.}}(\boldsymbol{x}_0, \theta) &= \sum_{l=2}^{L-1} \kappa(\text{SOPs}_l,\text{SOPs}_l^{\text{th}}) + \kappa(\boldsymbol{x}_{\text{L}},\boldsymbol{x}_{\text{L}}^{\text{th}})\\
\kappa(a, b) &= \frac{\max(0, a - b)}{b}\label{eq:loss_reg2}
\end{align}
where the superscript ``\(\text{th}\)'' denotes the threshold above which excess activity is penalized. We use $\text{SOPs}_l^{\text{th}}$ = 20M and $\boldsymbol{x}_{\text{L}}^{\text{th}}$ = 83.3k. We do not regularize the first layer, as its activity depends solely on sensor settings and scene dynamics, nor the off-chip gated decoding. The final layer's output spikes $\boldsymbol{x}_{\text{L}}$ are directly regularized to control readout load.

\begin{figure}[t]
	\centering
	\includegraphics[width=\linewidth]{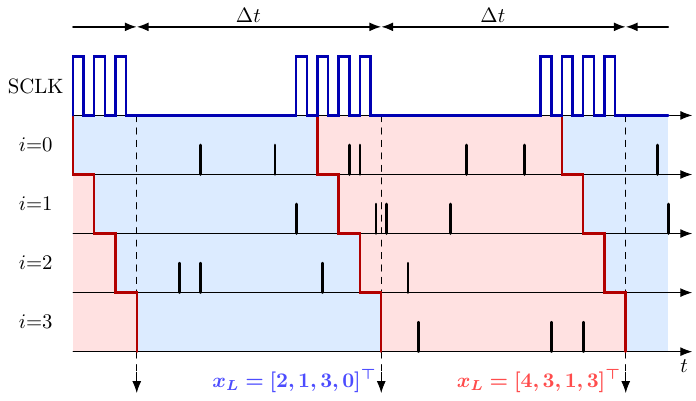}
    \vspace{-15pt}
    \caption{Cyclic readout strategy for a four-neuron layer. The MCU sequentially samples each neuron via SPI, synchronized to the SCLK signal, as rapidly as possible. Vertical black lines indicate spike events; colored regions show sampling windows. Dashed arrows mark the completion of each readout cycle, triggering the off-chip decoding process.
    }
	\label{fig:cyclic}
\end{figure}

\textbf{Optimizer and other details:} During training, event-camera sequences are processed sequentially in batches of 32, with a discretization window of $\Delta t$ = 10 ms. The loss is computed at each simulated timestep, and backpropagation is used to update the model parameters. We use the AdamW optimizer \cite{loshchilovdecoupled} with a weight decay of 0.05, as narrower weight distributions were empirically found to reduce the reality gap when deploying on the Speck2f SoC. Quantization-aware training is omitted as it provided negligible benefit for on-chip inference; instead, post-training quantization to 8-bit weights and 16-bit activations is applied before deployment.

\subsection{The road to an integrated solution}\label{sec:deployment}
Beyond our SNN and training procedure, additional mechanisms are required for a fully integrated solution where a low-power MCU interfaces two Speck2f devices via SPI. For more implementation-level details, including SPI commands structures, configuration encoding, and initialization steps, please refer to the supplementary material.

\textbf{Cyclic readout:} 
Retrieving the spiking activity from all output neurons over a time interval $\Delta t$ is a crucial preliminary step for our off-chip decoding process. However, as described in \secref{sec:speck}, a key challenge is the Speck2f readout core’s limitation to 16 accessible neurons, which must be polled sequentially via SPI. We address this with a cyclic readout strategy (see \figref{fig:cyclic}): the MCU time-multiplexes the sampling of each neuron’s activity within every $\Delta t$ window, synchronized to the SCLK signal, as rapidly as possible. This introduces minimal temporal offsets between readouts, mainly determined by the SCLK cycle duration (approx.\ 175 $\mu$s in our case). By cycling through all neuron indices, we construct output vectors $\boldsymbol{x}_L$ representing the final layer’s spiking activity for off-chip decoding. 

\begin{figure}[t]
	\centering
	\includegraphics[width=0.75\linewidth]{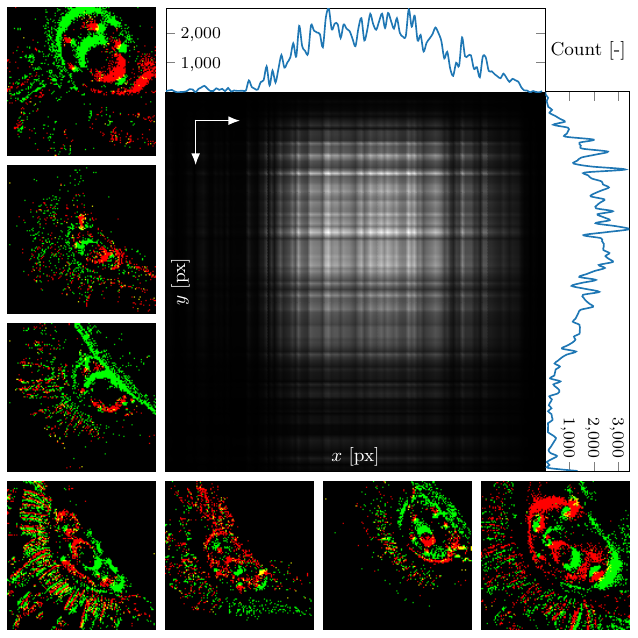}
    \vspace{-5pt}
    \caption{Representative samples from our real-world dataset (\textit{left} and \textit{bottom}) show event streams from different users and acquisition conditions. Colors used to distinguish event polarities: red for negative, green for positive. The \textit{central} heatmap displays the distribution of ground-truth pupil centers, with marginal histograms indicating coverage along each axis.}
	\label{fig:dataset}
\end{figure}

\begin{figure*}[t]
	\centering
	\includegraphics[width=0.975\linewidth]{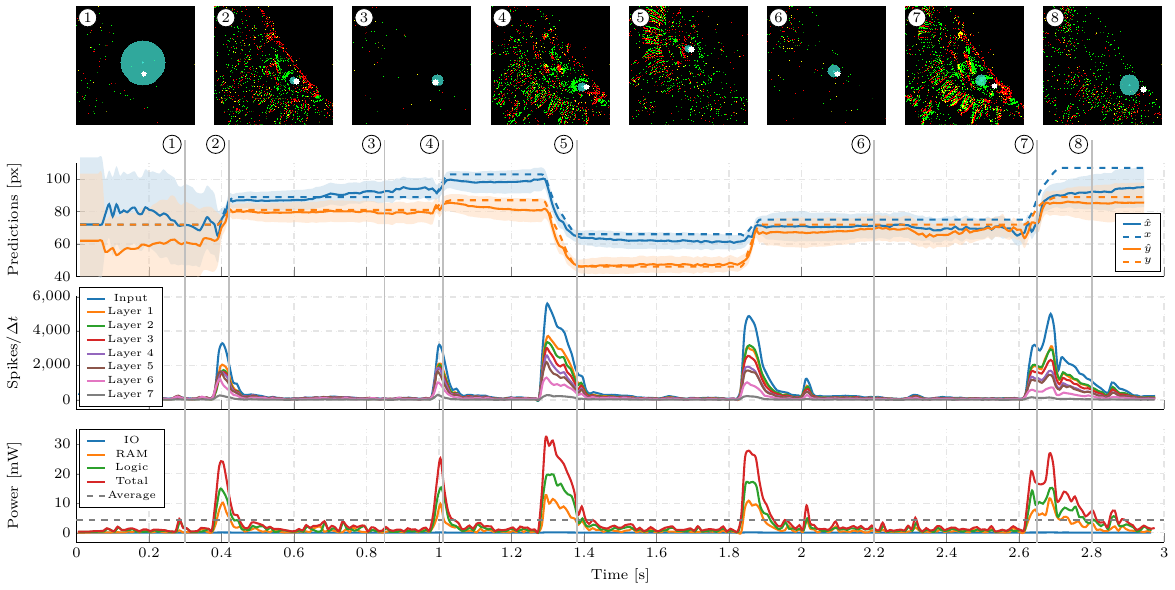}
    \vspace{-5pt}
    \caption{Example sequence from our real-world dataset and on-chip model evaluation. \textit{Top}: Event frames at key time points (white: ground truth; blue: prediction). \textit{Middle}: Predicted pupil coordinates with uncertainty, event counts, and spikes per layer. \textit{Bottom}: Processing power consumption for each Speck2f subsystem, with a total average of 4.22 mW. Vertical lines indicate correspondence between frames and time-series data. Results obtained with the Speck2f development kit.}
	\label{fig:activitypower}
\end{figure*}

\textbf{Spike multiplier:} A notable limitation of the Speck2f's readout core is that, as in Section~\ref{sec:speck}, it provides quantized moving averages over 16 SCLK cycles\footnote{Available moving-average windows are 1, 16, or 32 SCLK cycles~\cite{samna_docs}. We use 16 to retrieve output spikes over \(\Delta t\) in a single SPI transaction.}, rather than total spike counts. To bypass this moving average, simply increasing spike rates by scaling weights and thresholds is infeasible, as weights must remain below thresholds (see \secref{sec:arch}). To overcome this, we introduce an intermediate layer connecting each output neuron to $N$ neurons with unit weights and thresholds, effectively multiplying the spike count by $N$ at the cost of an additional core. We use \( N=4 \) instead of \( 16 \) due to a known hardware issue in mapping neurons to the readout core~\cite{samna_docs}, which necessitates adding a secondary layer with 4:1 connectivity (see supplementary material), thus effectively multiplying output activity by 16.

\subsection{Wearable hardware overview}
To realize a fully integrated pupil-center-tracking solution, we developed a custom printed circuit board (PCB) integrating two Speck2f devices (each with a 1.98 mm lens) and an nRF52840 MCU\footnote{MCU selection optimization for lower system-level power consumption was beyond the scope of this work.}, interfaced via SPI at 100 Hz. The MCU manages power, device programming, real-time Bluetooth monitoring, and output decoding. The system supports USB or dual 3.7 V lithium coin battery power, enabling untethered wearable use. For robust detection under varying lighting, the PCB includes two infrared (IR) LED rings with six LEDs each\footnote{Illumination optimization was beyond the scope of this work.}. Dedicated analog circuits monitor real-time power consumption of both Speck2f devices and the MCU. See \figref{fig:boards} (right) for the custom PCB and \figref{fig:combined_glasses_arch} (top) for the wearable prototype.

\section{Experiments}
Our primary objective is to achieve continuous pupil-center tracking with a fully integrated, low-power neuromorphic system. We do not directly compare to existing event-based methods, as prior work is generally unconstrained by hardware limitations and does not face the strict memory and bandwidth constraints of our system. Instead, we collected a dedicated real-world, multi-user dataset with the Speck2f for both training and validation, enabling rigorous evaluation under realistic operating conditions, which is then followed by an assessment using our wearable prototype.

\begin{figure*}[t]
	\centering
	\includegraphics[width=0.9675\linewidth]{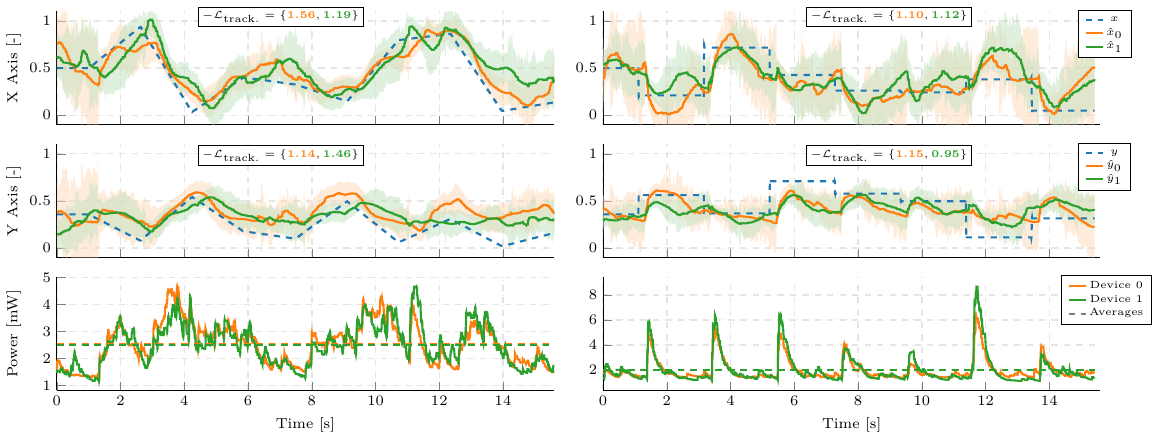}
    \vspace{-5pt}
    \caption{Wearable prototype results as a user performs smooth pursuit (left) and saccadic eye movements (right) while tracking a moving dot on a display at a fixed distance. \textit{Top and middle rows}: Ground-truth and predicted pupil-center coordinates with uncertainty estimates for left (device 0) and right (1) eyes. \textit{Bottom row}: Instantaneous power consumption per device and sequence averages. In both sequences, the MCU maintains an average power consumption of approx.\ 13.6 mW. Additional results can be found in the supplementary material.}
	\label{fig:v2time}
\end{figure*}

\subsection{Real-world dataset}
Our dataset\footnote{To be made publicly available upon publication.} contains 432 sequences from 8 users: 369 for training and 63 for validation. Each 3-second sequence captures data from a single eye as participants tracked a moving dot on a display at a fixed distance using a chinrest. To increase pupil size diversity, three background colors were used, and data was collected with four Speck2f devices to account for hardware variability. Natural eye behaviors, including blinks, were retained. Ground-truth pupil centers were manually annotated at 100~Hz. \figref{fig:dataset} shows representative samples and statistics. As shown, the Speck2f was angled to maximize eye region coverage. During training, data augmentation included random affine transformations.

\subsection{Results on real-world dataset}
We first evaluate our approach on the validation set of the real-world dataset, focusing on tracking accuracy and the reality gap. These experiments utilize the Speck2f development kit (see \figref{fig:boards}, left), which enables controlled event stream input and the monitoring of the internal activity of each layer. Quantitative results are presented in \tabref{tab:quant}, with qualitative trends illustrated in \figreftwo{fig:activitypower}{fig:sim2real}.

Our gated decoding configurations achieve the best on-chip performance. Models predicting shared and per-axis uncertainty yield average L2 errors (lower is better, $\downarrow$) of $9.99$ and $9.91$ pixels, respectively, closely matching their GPU results with minimal reality gap. Given their comparable accuracy, we select the shared uncertainty variant to reduce off-chip computation. Notably, models with uncertainty estimation achieve up to 24\% error reduction over no-uncertainty baselines. The performance advantage of gated over direct decoding (i.e., nearly 50\% error reduction across uncertainty configurations) underscores the importance of temporal reasoning for this task, suggesting that neuromorphic hardware supporting more sophisticated recurrent dynamics could enable further accuracy improvements.

\begin{table}[!t]
	\centering
	\resizebox{0.95\linewidth}{!}{%
	{\renewcommand{\arraystretch}{1.2} 
	\begin{tabular}{llccc}
		\thickhline
		\thickhline
        \multirow{2}{*}{\textbf{Decoding}} & \multirow{2}{*}{\textbf{Uncertainty}} & \multicolumn{2}{c}{\textbf{L2 Error [px]} $\boldsymbol{\downarrow}$} & \multirow{2}{*}{\textbf{SOPs [M/s]} $\boldsymbol{\downarrow}$}\\
		& & GPU & Speck2f & \\\thickhline
		\multirow{3}{*}{Direct} 
        & No       & 21.78 & $21.67 \pm 0.04$ & 8.73 \\
        & Shared   & 19.51 & $19.46 \pm 0.02$ & 15.33 \\
        & Per-axis & 19.24 & $19.08 \pm 0.01$ & 15.01 \\
        \cline{2-5}
		\multirow{3}{*}{Gated} 
        & No       & 13.11 & $13.24 \pm 0.01$ & \underline{5.61} \\
        & Shared   & \textbf{9.74} & \underline{$9.99 \pm 0.04$} & \textbf{5.57}\\
        & Per-axis & \underline{9.80} & $\boldsymbol{9.91 \pm 0.02}$ & 6.04 \\
		\thickhline
		\thickhline
	\end{tabular}}}
	\caption{Quantitative evaluation on our real-world dataset. Best in bold, runner up underlined. On-chip results are averaged over 5 runs to account for hardware stochasticity \cite{richter2023speck, yao2024spike}. Results obtained with the Speck2f development kit.}
	\label{tab:quant}
\end{table}

\begin{figure}[t]
	\centering
	\includegraphics[width=\linewidth]{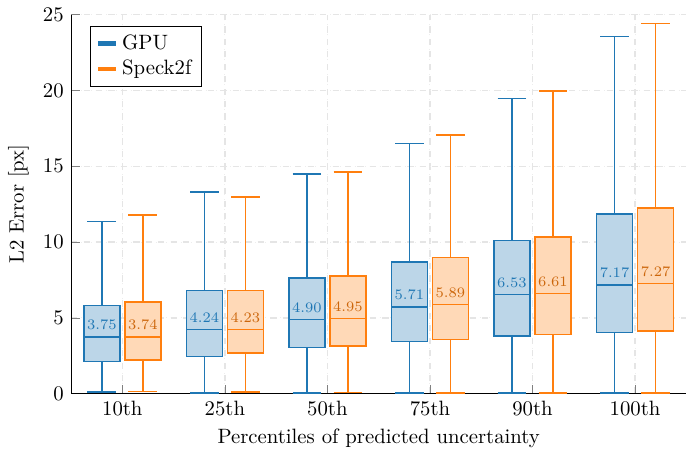}
    \caption{L2 error of the SNN model with gated decoding across percentiles of the shared predicted uncertainty for GPU-based and Speck2f-based (with development kit) evaluations.}
	\label{fig:sim2real}
\end{figure}

The selected model produces well-calibrated confidence estimates, evidenced by the correlation between predicted uncertainty and actual error in \figref{fig:sim2real}. On-chip median error increases from 3.74 pixels for the 10\% most confident predictions to 7.27 pixels across all samples, showing that the predicted uncertainty effectively identifies challenging cases. See the supplementary material for additional results.

Finally, \figref{fig:activitypower} highlights the energy efficiency and tracking performance of our solution over time. As shown, our predictions closely follow the ground-truth locations, demonstrating high on-chip tracking accuracy. Furthermore, the Speck2f maintains consistently low power consumption, averaging 4.22 mW, which strongly correlates with both the input event rate and internal spike activity.

\begin{figure}[t]
	\centering
	\includegraphics[width=\linewidth]{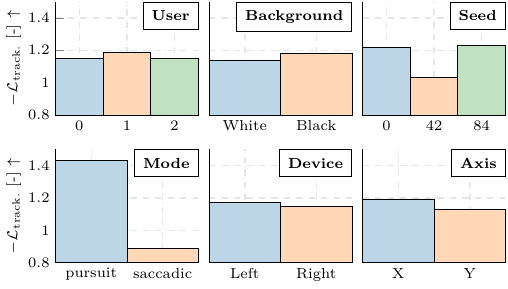}
    \caption{Tracking performance variability across conditions using our wearable prototype. Each subplot shows results for a specific condition, averaged over all other variables.}
	\label{fig:v2corr}
\end{figure}

\subsection{Results with wearable prototype}\label{sec:v2}
We demonstrate and evaluate our solution on our wearable prototype (see \figref{fig:combined_glasses_arch}, top), focusing on tracking accuracy and real-time, low-power operation. Three users wore the prototype and tracked a moving stimulus on a display at a fixed distance using a chinrest. To align the predicted and ground-truth pupil locations, we recorded a calibration sequence for each user, fitting an affine transformation via least squares. We evaluate performance across six factors: user identity, background color, stimulus seed (controlling trajectory variation), tracking mode, device (i.e., left/right eye), and coordinate axis, reporting accuracy via the negative mean tracking error $-\mathcal{L}_{\text{track.}}$ (higher is better, $\uparrow$). Quantitative results are shown in \figref{fig:v2corr}, with qualitative trends illustrated in \figref{fig:v2time}.

Our solution maintains consistent tracking performance across most conditions, demonstrating robust operation across users, backgrounds, devices, and axes (with slightly better performance along the x-axis), as shown in \figref{fig:v2corr}. The greatest variability appears in the stimulus seed and tracking mode. Different trajectory seeds produce varying difficulty levels based on speed, direction, and transition combinations. Additionally, saccadic movements yield considerably lower accuracy than smooth pursuit due to fundamental differences in eye movement dynamics and their interaction with event-based sensing. Smooth pursuit involves continuous, predictable tracking, generating sustained, high signal-to-noise ratio event streams. In contrast, saccades are rapid, ballistic eye movements producing bursty event patterns during transitions followed by minimal activity during fixations. This temporal sparsity challenges our architecture's limited temporal modeling capacity.

\figref{fig:v2time} illustrates these differences. During smooth pursuit (left), both eyes track closely with predicted coordinates following ground truth consistently and tight uncertainty estimates reflecting model confidence. Power consumption remains stable at around 2.5~mW per device. During saccadic movements (right), predicted trajectories show larger overall deviations from ground truth, with both uncertainty and errors increasing during the challenging fixation phases.  Power consumption exhibits brief spikes of 6--8~mW during saccades, corresponding to event bursts from rapid eye movements, before returning to a baseline during fixations of approx.\ 1.5 mW. Together with \figref{fig:activitypower}, these power profiles validate the event-driven nature of our system: power scales with visual dynamics, demonstrating efficient energy usage that responds directly to eye movement complexity.

\section{Limitations}
While our approach demonstrates fully integrated, low-power, event-based pupil tracking, several opportunities for improvement remain. First, our model employs simple IF neurons and a basic CNN structure due to Speck2f's strict constraints. More sophisticated architectures could enhance temporal reasoning but would require more flexible neuromorphic hardware. Similarly, the proposed spike multiplier mechanism consumes two of the nine processing cores to bypass readout limitations, reducing resources for network complexity. Future neuromorphic hardware with direct spike count retrieval or a dedicated core for more flexible on-chip decoding would eliminate this overhead.

Second, our current 100~Hz operating frequency matches our ground-truth annotation rate. This represents an underutilized system capability: the sparse, asynchronous nature of the SNN enables scaling to higher frequencies with negligible power increase, as processing remains event-driven and proportional to visual dynamics rather than output sampling rate. Given access to high-frequency ground truth for decoder training, our system could achieve significantly higher tracking rates with minimal additional power cost; a key scalability advantage of our neuromorphic approach.

Finally, the reported tracking errors stem from hardware constraints and dataset complexity, including multi-user variability, blinks, and rapid movements followed by long fixations. Despite these limitations, our solution suits applications requiring approximate gaze estimation, such as vergence estimation or coarse gaze-based interaction.

\section{Conclusion}
This work bridges the gap between neuromorphic hardware capabilities and practical deployment requirements for continuous wearable eye tracking. By addressing the unique challenges of tight integration, we demonstrate that this technology can power practical, always-on event-based vision systems with minimal energy overhead. Our SNN, powered by the gated decoding and uncertainty predictions, enables robust, continuous pupil-center tracking while respecting the stringent constraints of on-chip processing. The resulting fully integrated system, featuring two Speck2f SoCs, achieves sub-5 mW average power consumption per eye without sacrificing accuracy, demonstrating that neuromorphic processors can deliver practical performance beyond laboratory prototypes. Beyond eye tracking, the principles and methods introduced here provide a blueprint for intelligent, energy-efficient systems across diverse event-based computer vision tasks requiring always-on operation.

\section*{Acknowledgments}
We thank Lyes Khacef for his valuable feedback and support throughout this work. We are also grateful to SynSense for their technical support, which was essential for the successful deployment of our models on the Speck2f hardware.

{
    \small
    \bibliographystyle{ieeenat_fullname}
    \bibliography{main}
}

\clearpage

\twocolumn[
\begin{center}
    {\Large\bf Realizing Fully-Integrated, Low-Power, Event-Based\\\vspace{4pt}Pupil Tracking with Neuromorphic Hardware}\\
    \vspace{5pt}
    {\large\bf --- Supplementary Material ---}\\
    \vspace{20pt}
    {\large Federico Paredes-Vall\'es$^{1}$\hspace{16pt}Yoshitaka Miyatani$^{2}$\hspace{16pt}Kirk Y. W. Scheper$^1$\vspace{5pt}\\
   $^{1}$Sony Advanced Visual Sensing AG, Schlieren, Switzerland\\
   $^{2}$Sony Semiconductor Solutions Corporation, Atsugi, Japan}
\end{center}
\vspace{20pt}
]

\renewcommand{\thesection}{S\arabic{section}}
\renewcommand{\thesubsection}{S\arabic{section}.\arabic{subsection}}
\renewcommand{\thesubsubsection}{S\arabic{section}.\arabic{subsection}.\arabic{subsubsection}}
\renewcommand{\thetable}{S\arabic{table}}
\setcounter{table}{0}
\setcounter{section}{0}

\section{Uncertainty calibration}
To validate our   uncertainty estimates~\cite{kendall2017uncertainties}, we analyze its calibration using reliability diagrams adapted for regression. For each prediction $\hat{\boldsymbol{y}}$ of pupil center location with estimated standard deviations $\hat{\boldsymbol{\sigma}} = (\hat{\sigma}_x, \hat{\sigma}_y)$, we compute the normalized error via the Mahalanobis distance:
\begin{equation}    
z = \sqrt{(\boldsymbol{y} - \hat{\boldsymbol{y}})^\top \boldsymbol{\Sigma}^{-1} (\boldsymbol{y} - \hat{\boldsymbol{y}})},
\end{equation}
where $\boldsymbol{\Sigma} = \text{diag}(\hat{\sigma}_x^2, \hat{\sigma}_y^2)$ is the diagonal covariance matrix.

Under a bivariate Gaussian assumption with diagonal covariance, $z^2$ follows a $\chi^2$ distribution with two degrees of freedom. We convert each $z$ to its corresponding cumulative probability under this distribution and plot the empirical frequency with which these probabilities fall below given thresholds (y-axis) against the theoretical probabilities (x-axis). Perfect calibration corresponds to the diagonal $x=y$; curves above indicate underconfidence (i.e., overestimated uncertainties), while curves below indicate overconfidence. 

\figref{fig:calibration} presents calibration curves for our main model variants, with calibration quality quantified by the mean squared error (MSE) from the diagonal. As shown, direct decoding models achieve superior calibration as uncertainty is their sole mechanism for handling low signal-to-noise periods, such as during eye fixations or blinks. In contrast, gated decoding models can also rely on gating for such periods, introducing slight redundancy that degrades uncertainty calibration. However, gated models achieve substantially lower L2 errors (see Table~1, main paper), motivating their adoption for deployment.

All models exhibit mild underconfidence, with predicted uncertainty intervals being wider than the observed error distribution warrants. We consider this conservative behavior preferable to overconfidence: it ensures that stated confidence intervals reliably contain the true pupil location, albeit at the cost of slightly overestimating the interval width. This provides trustworthy uncertainty bounds for potential downstream applications, where underestimating positional uncertainty could compromise system reliability.

\renewcommand{\thefigure}{S1}
\begin{figure}[t]
	\centering
	\includegraphics[width=\linewidth]{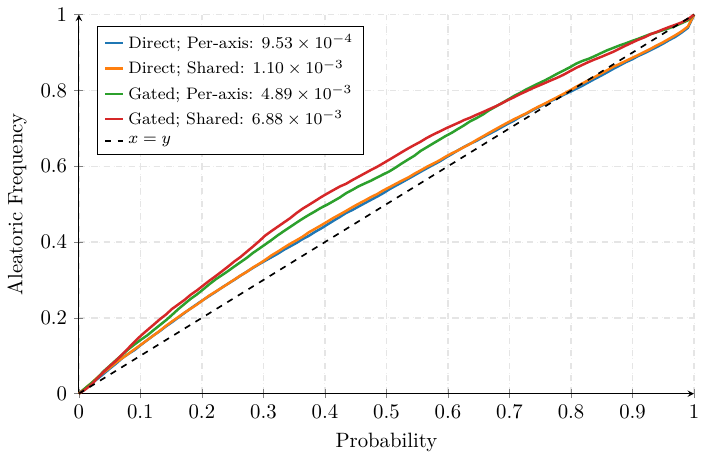}
\caption{Reliability diagrams for the main model variants. Empirical cumulative frequencies are plotted against theoretical probabilities under a bivariate Gaussian assumption. MSE values quantify calibration quality relative to the diagonal (perfect calibration).}
	\label{fig:calibration}
\end{figure}

\renewcommand{\thefigure}{S2}
\begin{figure}[t]
	\centering
	\includegraphics[width=\linewidth]{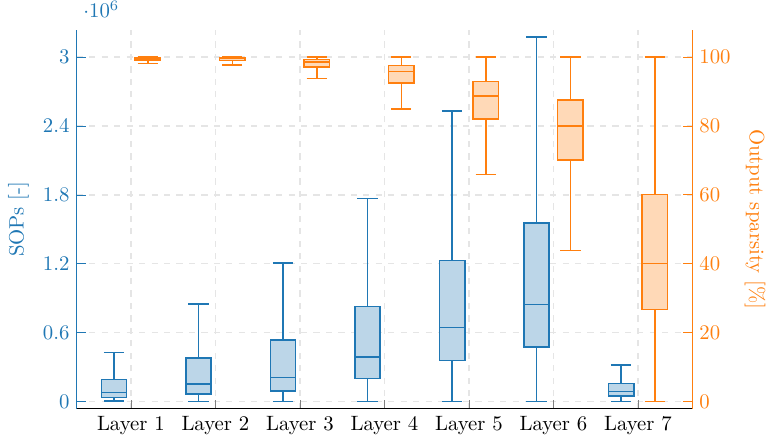}
    \caption{SOPs and output sparsity per layer for the SNN model with gated decoding and shared uncertainty estimation, computed over $\Delta t = 10$ ms windows in simulation.}
	\label{fig:sops}
\end{figure}

\renewcommand{\thefigure}{S3}
\begin{figure*}[t]
	\centering
	\includegraphics[width=\linewidth]{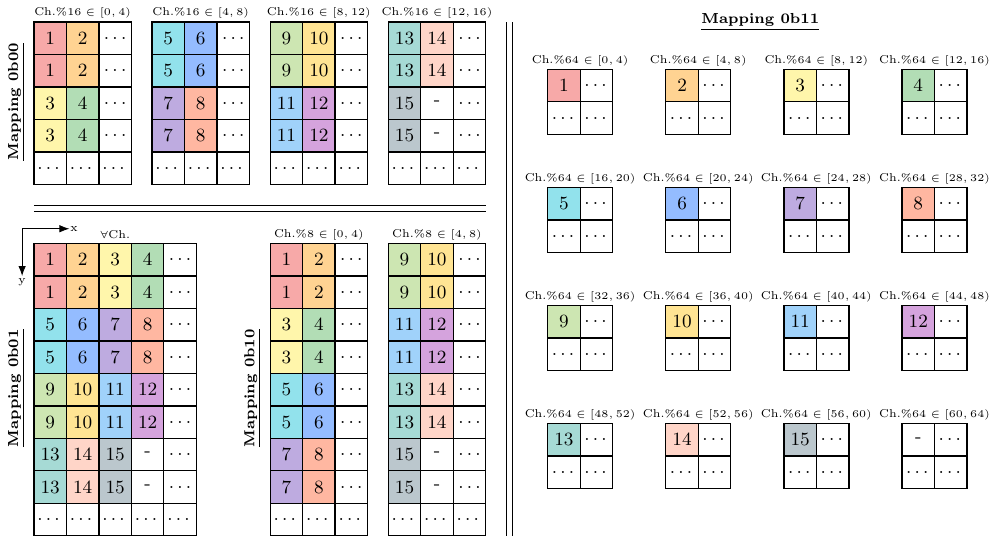}
    \caption{Readout core address mapping characterization. Numbers indicate readout neuron assignments for each feature map location. Repeated numbers denote address collisions; dashes mark inaccessible positions. Ellipsis notation indicates repetition in the corresponding direction: horizontally if aligned right, vertically if aligned below, or diagonally if positioned diagonally.}
	\label{fig:hardwareissue}
\end{figure*}

\section{Computational load and activation sparsity}
\figref{fig:sops} characterizes the computational load and activity patterns across spiking layers on our real-world validation dataset in simulation. Results were obtained with our SNN featuring gated decoding and shared aleatoric uncertainty predictions. The left y-axis (blue) shows SOPs, while the right y-axis (orange) presents layer sparsity, defined as the percentage of spiking neurons remaining silent, both computed using $\Delta t = 10$ ms time windows.

\renewcommand{\thefigure}{S4}
\begin{figure*}[t]
	\centering
	\includegraphics[width=0.985\linewidth]{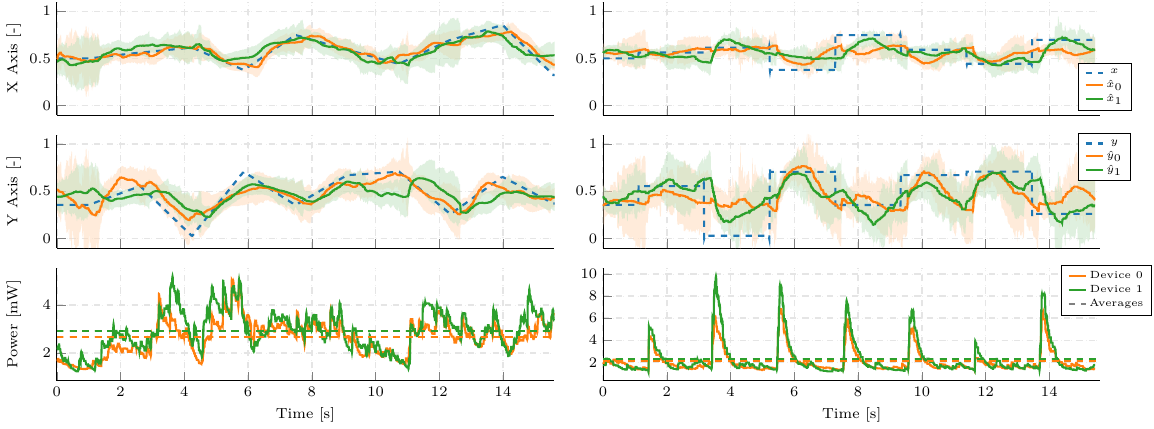}\\\vspace{5pt}
    \includegraphics[width=0.985\linewidth]{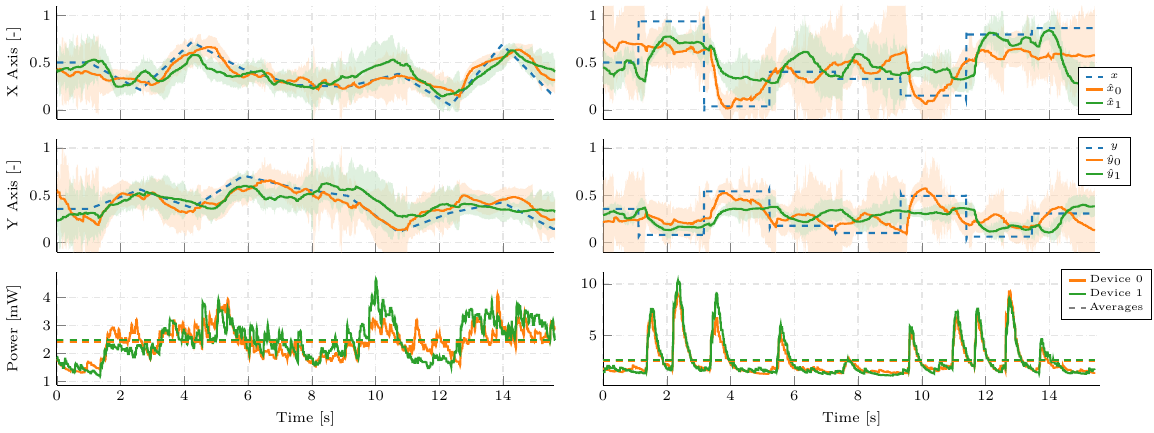}
    \caption{Additional wearable prototype results as a user performs smooth pursuit (left) and saccadic eye movements (right) while tracking a moving dot on a screen at a fixed distance. \textit{Top and middle rows}: Ground-truth and predicted pupil-center coordinates with uncertainty estimates for left (device 0) and right (1) eyes. \textit{Bottom row}: Instantaneous power consumption per device and sequence averages.}
	\label{fig:v2time2}
\end{figure*}

The SOPs distribution reveals that processing demands scale with layer depth, increasing from a median of approx.\ 0.1M in Layer 1 to 0.8M in Layer 6 before dropping sharply in the final layer. This trend reflects the network's hierarchical feature processing: early layers operate on high-resolution, sparse event inputs, while deeper layers process increasingly abstract representations with denser activation patterns. The dramatic reduction in Layer 7 stems from its role as the output layer with only 15 neurons, minimizing both computational cost and readout bandwidth. All layers operate well below their respective bandwidth thresholds (i.e., 100M SOPs for Layer 1, 30M for Layers 2--7 \cite{yao2024spike}), validating our activity regularization strategy from Section 3.4 of the main paper.

Sparsity patterns show complementary behavior. Layers 1--3 maintain extremely high sparsity, with most neurons remaining inactive due to the sparse nature of event-based inputs and the feature map resolution. Sparsity gradually decreases through Layers 4--6 as features become more abstract and activation patterns more distributed. Layer 7 exhibits the highest variability (median approx.\ 40\%, range 0--100\%), reflecting its role in encoding both pupil location and aleatoric uncertainty: high sparsity during periods of minimal eye movement or occlusions, and dense activity during active tracking. This adaptive behavior demonstrates that our gated decoding mechanism effectively modulates output activity based on input dynamics, contributing to the event-driven power efficiency shown in Figs.\ 5 and 6 of the main paper.

\section{Speck2f's hardware issue}
Speck2f's readout core contains a hardware errata \cite{samna_docs} affecting neuron addressing when mapping feature maps to its 16 monitoring neurons. Of these, the 0th neuron remains always silent \cite{samna_docs}, leaving 15 accessible neurons for readout. This core nominally supports multiple spatial configurations, in [channel, height, width] format:

\setlist[itemize]{
    leftmargin=20pt,
    itemsep=2pt,
    topsep=4pt,
    parsep=0pt
}
\begin{itemize}
    \item Mapping \texttt{0b00} -- Designed for [4, 2, 2] maps.
    \item Mapping \texttt{0b01} -- Designed for [2, 2, 4] maps.
    \item Mapping \texttt{0b10} -- Designed for [1, 4, 4] maps.
    \item Mapping \texttt{0b11} -- Designed for [16, 1, 1] maps.
\end{itemize}

However, the addressing logic incorrectly resolves spatial coordinates in all configurations. To characterize this hardware errata, we conducted a systematic probing experiment where we stimulated individual neurons across spatial positions and channel indices while monitoring which readout neurons received spikes. \figref{fig:hardwareissue} illustrates the resulting address mapping behavior for each configuration. Our characterization reveals systematic address duplication with configuration-dependent patterns:

\begin{itemize}
    \item Mapping \texttt{0b00}: Addresses repeat vertically in pairs and every four positions, horizontally every two positions, and across four-channel groups, requiring a minimal $[13, 4, 2]$ feature map to activate the 15 readout neurons.
    \item Mapping \texttt{0b01}: Addresses repeat vertically in pairs and every eight positions, horizontally every four positions, and across all channels, requiring a minimal $[1, 8, 4]$ feature map to activate the 15 readout neurons.
    \item Mapping \texttt{0b10}: Addresses repeat vertically in pairs and every eight positions, horizontally every two positions, and across four-channel groups, requiring a minimal $[5, 8, 2]$ feature map to activate the 15 readout neurons.
    \item Mapping \texttt{0b11}: Addresses repeat across four-channel groups only, requiring a minimal $[57, 1, 1]$ feature map to activate the 15 readout neurons.
\end{itemize}

\renewcommand{\thefigure}{S5}
\begin{figure}[t]
	\centering
	\includegraphics[width=0.65\linewidth]{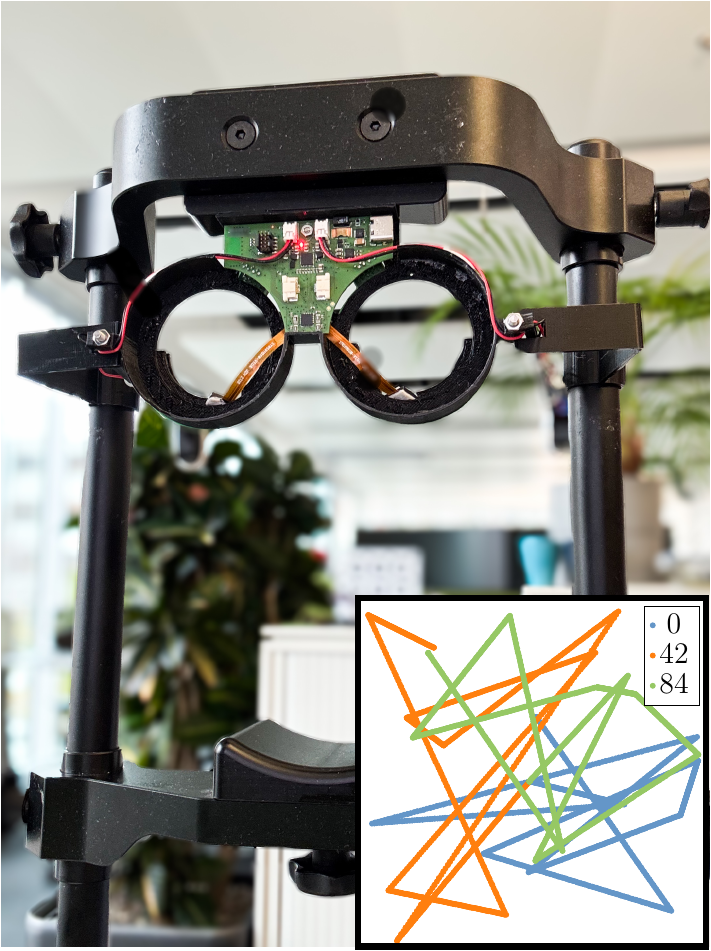}
    \caption{Our wearable prototype mounted on a chinrest for controlled data collection. \textit{Inset} shows stimulus trajectories used for both smooth pursuit and saccadic eye movement experiments. Legend indicates random seed values for trajectory generation.}
	\label{fig:setupstimulus}
\end{figure}

Our core SNN architecture must route 15 output neurons to the readout core while amplifying their spike counts by a factor of 16 to compensate for the quantized moving average (see Section 3.5, main paper). Based on the results of our hardware characterization, we adopt mapping configuration \texttt{0b11} preceded by a $[60, 1, 1]$ layer. This configuration exploits the four-channel grouping pattern inherent to the errata: each group of four consecutive channels (e.g., channels 0--3, 4--7, etc.) maps to the same readout neuron, effectively providing a $4\times$ amplification. This, appended to our proposed spike multiplier (see Section 3.5, main paper) which already contributes a $4\times$ amplification, achieves the required $16\times$ increase in output activity.

\section{Additional results with wearable prototype}
We provide additional wearable prototype results to complement the main paper. \figref{fig:v2time2} shows representative tracking sequences under various conditions, reinforcing the conclusions from Section~4.3 of the main paper. Across all sequences, we observe consistent performance during smooth pursuit with tight uncertainty bounds and accurate trajectory following, while power consumption remains stable at approx.\ 2.5~mW per Speck2f device. During saccadic movements, we consistently observe larger overall errors with uncertainty peaking during fixations due to the sparse input events, alongside characteristic power spikes of approx.\ 5--10~mW corresponding to event bursts. The physical experimental setup and stimulus trajectory patterns used for these experiments are shown in \figref{fig:setupstimulus}.

\subsection{Supplementary video}

Alongside this document, we provide a video\footnote{The video is available at \url{https://bit.ly/4io51iL}} demonstrating real-time operation of our wearable prototype. The video shows live tracking performance, uncertainty estimation, and dynamic power scaling during natural eye movements. We encourage readers to watch it for a complete understanding of our system's capabilities. A snapshot of the real-time monitoring interface is shown in \figref{fig:gui}.

\renewcommand{\thefigure}{S6}
\begin{figure}[t]
	\centering
	\includegraphics[width=\linewidth, trim=0 70pt 0 0, clip]{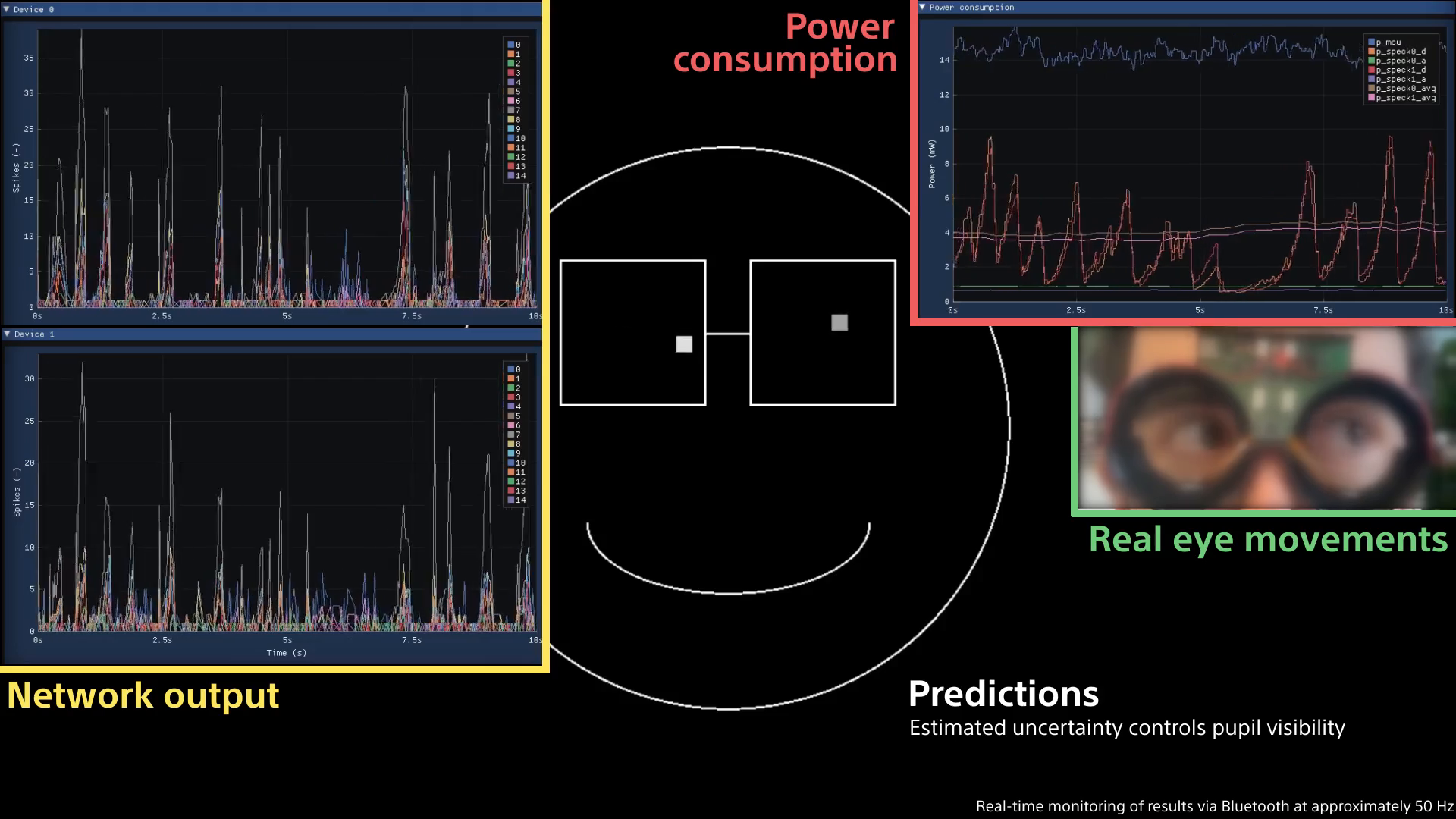}
     \caption{Real-time monitoring interface for the wearable prototype. The interface displays predicted pupil center locations with uncertainty estimates (\textit{center}), power consumption of individual Speck2f devices and MCU (\textit{top right}), and raw SNN output activity (\textit{left}). Data is streamed via Bluetooth at approx.\ 50~Hz. \textit{Inset} shows corresponding eye movements captured by an external camera for validation (blurred for anonymity).}
	\label{fig:gui}
\end{figure}

\section{Speck2f operation}
This section describes the communication protocols and initialization procedures for interfacing with Speck2f without the its FPGA-based development kit. These details complement the official Speck2f documentation~\cite{speck_specsheet, samna_docs} to enable replication of our setup.

\subsection{SPI communication}\label{sec:spi_comm}
For each register access, a three-byte command header is constructed prior to SPI transmission to specify the operation type and target register address. The first byte encodes the access mode, where the most significant bit is set to~1 for write operations and cleared for read operations. The following two bytes represent the 16-bit target register address, transmitted in big-endian order. After this header, any associated payload bytes (i.e., data to be written or space reserved for incoming data in a read operation) are transmitted sequentially as part of the same SPI frame.

\subsection{Reformatting the configuration file}
This section presents a systematic approach for compressing and reformatting the binary configuration file generated by SynSense's Samna API~\cite{samna_docs} for programming the Speck2f via custom SPI-based communication. Tables~\ref{tab:constants} and~\ref{tab:config_info} summarize important constants and addressing information.

\begin{table}[!t]
	\centering
	\resizebox{0.75\linewidth}{!}{%
	{\renewcommand{\arraystretch}{1.2} 
	\begin{tabular}{ll}
		\thickhline
		\thickhline
        \textbf{Constant} & \textbf{Value} \\\thickhline
		KERNEL\_START & 0x4C0\\
        R\_START & 0x100\\
        KERNEL & [0x10000, 0x14000, 0x18000,\\
         & 0x20000, 0x28000, 0x30000,\\
         & 0x40000, 0x50000, 0x54000]\\
         CNN\_REG & [0x280, 0x2C0, 0x300, 0x340\\
         &  0x380, 0x3C0, 0x400, 0x440,\\
          &  0x480]\\
		\thickhline
		\thickhline
	\end{tabular}}}
	\caption{Constants and addresses required during the preparation phase of the Speck2f configuration file.}
	\label{tab:constants}
\end{table}

\textbf{Kernel memory size computation:} For each layer $l$, we first extract configuration parameters from the original binary file (denoted ``$\text{binary}$''). For the kernel size:
\begin{equation}
k_s(l) = (\text{binary}[\text{CNN\_REG}[l] + 0\text{x1}] \gg 2) \,\&\, 0\text{xF}
\end{equation}
where $\gg$ and $\&$ denote bitwise right-shift and AND.

The output feature count is reconstructed from its low and high byte components:
\begin{align}
C_{\text{out\_low}}(l) &= \text{binary}[\text{CNN\_REG}[l] + \text{0x04}] \gg 6\\
C_{\text{out\_high}}(l) &= \text{binary}[\text{CNN\_REG}[l] + \text{0x05}]\\
C_{\text{out}}(l) &= (C_{\text{out\_high}}(l) \ll 2) \,|\, C_{\text{out\_low}}(l)
\end{align}
where $\ll$ and $|$ denote bitwise left-shift and OR.

Similarly, the input feature count is obtained as:
\begin{align}
C_{\text{in\_low}}(l) &= \text{binary}[\text{CNN\_REG}[l] + \text{0x02}]\\
C_{\text{in\_high}}(l) &= \text{binary}[\text{CNN\_REG}[l] + \text{0x03}] \& \text{0b11}\\
C_{\text{in}}(l) &= (C_{\text{in\_high}}(l) \ll 8) \,|\, C_{\text{in\_low}}(l)
\end{align}

The kernel memory size is then computed as:
\begin{align}
k_{\text{mem}}(l) &= \left((k_s(l) + 1)^2 - 1\right) \ll (w_{\text{in}}(l) + w_{\text{out}}(l)) \nonumber\\
&\quad |\, (C_{\text{out}}(l) \ll w_{\text{in}}(l)) \,|\, C_{\text{in}}(l) + 1
\end{align}
where $w_{\text{in}}(l) = \lceil \log_2(C_{\text{in}}(l) + 1) \rceil$ and $w_{\text{out}}(l) = \lceil \log_2(C_{\text{out}}(l) + 1) \rceil$ are the required bit widths.

\textbf{File compression:} We run a compression process that iterates through the nine potentially active layers (i.e., $l = 0$ to $8$), extracting and packing configuration data. The compressed file begins with $\text{binary}[0:\text{KERNEL\_START}]$, preserving layer shapes and spiking thresholds. For each layer, we determine if it is active as follows:
\begin{align}
&\text{dest}(l) = \text{binary}[\text{R\_START} + \text{0x09} + \lfloor l/4 \rfloor]\\
&\text{layer\_active\_mask} = \text{binary}[\text{R\_START} + \text{0x27}]\\
&\text{is\_active}(l) = \left(\text{dest}(l) \gg (2(l \bmod 4))\right) \& \text{0b11} \nonumber\\
&\quad\quad\quad\quad\quad \lor\, \left(\text{layer\_active\_mask} \gg l\right) \& \text{0x01}
\end{align}
where $\lor$ denotes logical OR.

For active layers, we append: (1)~1-byte layer ID $l$, (2)~2-byte $k_{\text{mem}}(l)$ in little-endian format, (3)~$k_{\text{mem}}(l)$ bytes of kernel weights from $\text{binary}[\text{KERNEL}[l]]$, and (4)~2-byte zero for unused leak data. Inactive layers are represented as $[l, 0, 0, 0, 0]$.

\begin{table}[t]
    \centering
    \begin{minipage}[b]{0.3275\textwidth}
        \centering
        \resizebox{\textwidth}{!}{
            \begin{tabular}{ccc}
		\thickhline
		\thickhline
        \textbf{Reg. addr.} & \textbf{Config. addr.} & \textbf{Num. regs.} \\\thickhline
		0x0000 & 0x0040 & 34\\
        0x0600 & 0x0280 & 17\\
        0x0700 & 0x02C0 & 17\\
        0x0800 & 0x0300 & 17\\
        0x0900 & 0x0340 & 17\\
        0x0A00 & 0x0380 & 17\\
        0x0B00 & 0x03C0 & 17\\
        0x0C00 & 0x0400 & 17\\
        0x0D00 & 0x0440 & 17\\
        0x0E00 & 0x0480 & 17\\
        0x0100 & 0x0080 & 17\\
        0x0118 & 0x0098 & 6\\
        0x0300 & 0x0100 & 45\\
        0x0200 & 0x00C0 & 23\\
		\thickhline
		\thickhline
	\end{tabular}
        }
    \end{minipage}%
    \hspace{0.01\textwidth}
    \begin{minipage}[b]{0.1125\textwidth}
        \centering
        \resizebox{\textwidth}{!}{
            \begin{tabular}{c}
		\thickhline
		\thickhline
        \textbf{Kernel addr.}\\\thickhline
		0x220000\\
        0x240000\\
        0x260000\\
        0x280000\\
        0x2A0000\\
        0x2C0000\\
        0x2E0000\\
        0x300000\\
        0x320000\\
		\thickhline
		\thickhline
	\end{tabular}
        }
    \end{minipage}
    \caption{Register (left) and memory information (right) used in the configuration encoding process.}
    \label{tab:config_info}
\end{table}

\textbf{File encoding:} The compressed file is converted into SPI command streams for device programming through register and memory configuration operations.

The register configuration process extracts layer parameters from the compressed configuration file and prepares SPI write commands following the protocol in Section~\ref{sec:spi_comm}. For each entry $(a_{\text{reg}}, a_{\text{config}}, n)$ in Table~\ref{tab:config_info} (left), $n$ consecutive bytes are read from address $a_{\text{config}}$ and written to Speck2f registers $a_{\text{reg}}$ through $a_{\text{reg}} + n - 1$. Each register write is prepended with a 4-byte big-endian length prefix, enabling SPI controllers to parse transaction boundaries and sequentially transmit individual commands without delimiters or fixed packet sizes.

The memory configuration process prepares the transfer of kernel weights to the Speck2f's on-chip memory banks. For each active layer in the compressed file (starting at address KERNEL\_START), if $k_{\text{mem}}(l) > 0$, the kernel data is prepared for writing to memory address $\text{mem\_addr}[l]$ from Table~\ref{tab:config_info} (right). Each memory write requires a two-phase protocol. First, six registers are configured to specify the target memory address and transfer length:
\begin{align}
&\text{0x6} \leftarrow (\text{mem\_addr[l]} \gg 16) \& \text{0xFF} \nonumber\\
&\text{0x5} \leftarrow (\text{mem\_addr[l]} \gg 8) \& \text{0xFF} \nonumber\\
&\text{0x4} \leftarrow \text{mem\_addr[l]} \& \text{0xFF} \nonumber\\
&\text{0x3} \leftarrow (k_{\text{mem}}(l) \gg 16) \& \text{0xFF} \nonumber\\
&\text{0x2} \leftarrow (k_{\text{mem}}(l) \gg 8) \& \text{0xFF} \nonumber\\
&\text{0x1} \leftarrow k_{\text{mem}}(l) \& \text{0xFF}\nonumber
\end{align}
where, as before, each register write follows the standard SPI protocol plus a 4-byte big-endian length prefix. 

Second, the bulk data transfer is initiated by constructing a memory write transaction with length $3 + k_{\text{mem}}(l)$ (i.e., 3-byte command header plus payload). This length is encoded as a 4-byte big-endian prefix, and is followed by the command header [0xE0, 0x00, 0x00] that identifies a memory write operation, and finally the $k_{\text{mem}}(l)$ kernel weight bytes. This sequence is repeated for each active layer. The final encoded bytearray contains all register and memory write commands, each with its length prefix, ready for sequential SPI transmission.

\textbf{Device initialization:} After programming the configuration, the Speck2f requires initialization via a sequence of control register writes. The initialization sequence writes commands 0x01, 0x81, 0xC1, 0xE1, and 0xF1 to register address 0x0 following the SPI protocol in Section~\ref{sec:spi_comm}. These commands can be appended to the encoded configuration bytearray for automatic device startup after programming.

\subsection{Cyclic readout}
To monitor spike activity across all $16$ readout neurons, we propose a cyclic readout mechanism that sequentially accesses each neuron's spike counter (see Section 3.5, main paper). For each neuron, we toggle the SCLK GPIO pin to trigger a clock cycle, then perform three SPI transmissions: two read operations to retrieve the 16-bit spike count from registers 0x213 (low byte) and 0x214 (high byte), followed by a write to register 0x20B that configures the next neuron for monitoring. The next neuron index is encoded as ((\texttt{next\_neuron} \& 0xFF) $<<$ 2) $|$ 0x83 to ensure proper alignment and command formatting.

\end{document}